\documentclass[10pt,conference,a4paper]{IEEEtran}
\usepackage[top=0.78in, bottom=1.61in, left=0.53in, right=0.53in]{geometry}
\ifCLASSOPTIONcompsoc
  \usepackage[nocompress]{cite}
\else
  \usepackage{cite}
\fi
\ifCLASSINFOpdf
  \usepackage[pdftex]{graphicx}
\else
\fi
\usepackage{amsmath}
\usepackage{algorithmic}
\ifCLASSOPTIONcompsoc
 \usepackage[caption=false,font=footnotesize,labelfont=sf,textfont=sf]{subfig}
\else
 \usepackage[caption=false,font=footnotesize]{subfig}
\fi
\usepackage{url}

\usepackage{multirow}
\usepackage{bm}
\usepackage{algorithm}
\usepackage{mathrsfs}
\usepackage{epstopdf}
\usepackage{threeparttable}
\usepackage{bbding}
\usepackage{tabularx}
\usepackage{booktabs}  

\newcolumntype{I}{!{\vrule width 0.5pt}}
\newlength\savedwidth
\newcommand\whline{\noalign{\global\savedwidth\arrayrulewidth
                            \global\arrayrulewidth 1pt}%
                   \hline
                   \noalign{\global\arrayrulewidth\savedwidth}}
\newlength\savewidth
\newcommand\shline{\noalign{\global\savewidth\arrayrulewidth
                            \global\arrayrulewidth 0.5pt}%
                   \hline
                   \noalign{\global\arrayrulewidth\savewidth}}

\begin{document}

\title{Learning Spatial-Semantic Context with Fully Convolutional Recurrent Network for Online Handwritten Chinese Text Recognition}

\author{\IEEEauthorblockN{Zecheng Xie$^1$, Zenghui Sun$^1$, Lianwen Jin$^{1\ast}$, Hao Ni$^{24}$, Terry Lyons$^{34}$}
\IEEEauthorblockA{
$^1$College of Electronic and Information Engineering, South China University of Technology, Guangzhou, China\\
$^2$Oxford-Man Institute for Quantitative Finance, University of Oxford, Oxford, UK \\
$^3$Mathematical Institute, University of Oxford, Oxford, UK\\
$^4$The Alan Turing Institute, University of Oxford, Oxford, UK\\
xiezcheng@foxmail.com, sunfreding@gmail.com, $^\ast$lianwen.jin@gmail.com, hao.ni@maths.ox.ac.uk, tlyons@maths.ox.ac.uk}

}

\maketitle
\begin{abstract}
Online handwritten Chinese text recognition (OHCTR) is a challenging problem as it involves a large-scale character set, ambiguous segmentation, and variable-length input sequences. 
In this paper, we exploit the outstanding capability of path signature to translate online pen-tip trajectories into informative signature feature maps using a sliding window-based method, 
successfully capturing the analytic and geometric properties of pen strokes with strong local invariance and robustness.
A multi-spatial-context fully convolutional recurrent network (MC-FCRN) is proposed to exploit the multiple spatial contexts from the signature feature maps and generate a prediction sequence while completely avoiding the difficult segmentation problem.
Furthermore, an implicit language model is developed to make predictions based on semantic context within a predicting feature sequence, 
providing a new perspective for incorporating lexicon constraints and prior knowledge about a certain language in the recognition procedure.
Experiments on two standard benchmarks, Dataset-CASIA and Dataset-ICDAR, yielded outstanding results, with correct rates of 97.10\% and 97.15\%, respectively, which are significantly better than the best result reported thus far in the literature. 
\end{abstract}

\IEEEpeerreviewmaketitle

\section{Introduction}

%
%
%
%

 

\begin{figure*}[t]
\centering
\includegraphics[width=\textwidth]{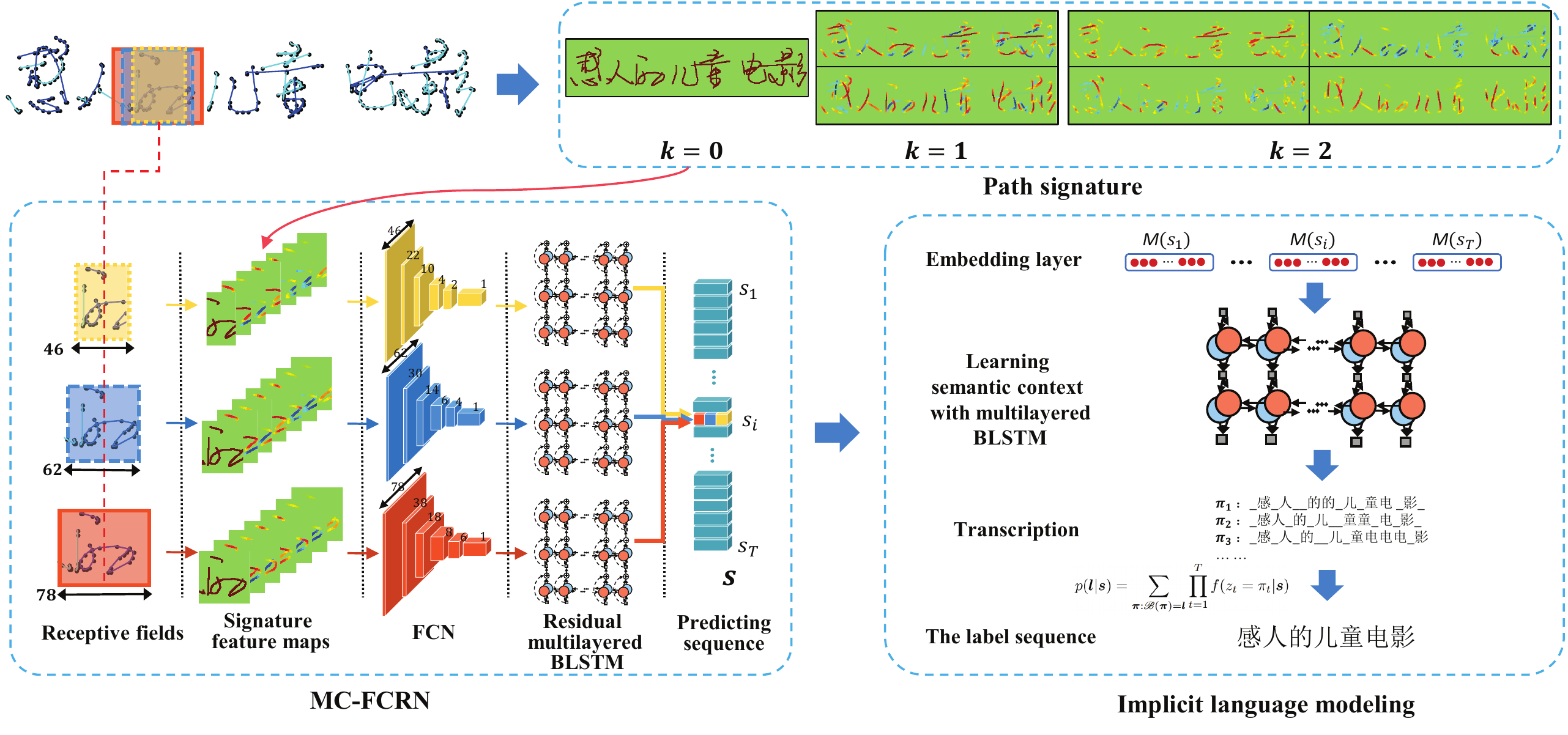}
\caption{Overview of the proposed method. Variable-length pen-tip trajectories are first translated into offline signature feature maps that preserve the essential online information. Then, a multi-spatial-context fully convolutional recurrent network (MC-FCRN) take input of the signature feature maps with receptive fields of different scales in a sliding window manner and generate a predicting sequence.
Finally, an implicit language model is proposed to derive the final label sequence by exploiting the semantic context of embedding vectors that are transformed from the predicting sequence.
}
\label{Fig:network}
\end{figure*}

In recent years, increasingly in-depth studies have led to significant developments in the field of handwritten text recognition. Various methods have been proposed by the research community, including integrated segmentation-recognition methods \cite{wang2012approach,wang2012handwritten,zhou2013handwritten,zhou2014minimum}, hidden Markov models (HMMs) and their hybrid variants \cite{bengio1999markovian,jiang2011novel,su2009off}, segmentation-free methods \cite{graves2009novel,graves2012supervised,liwicki2007novel,messina2015segmentation} with long short-term memory (LSTM) and multi-dimensional long short-term memory (MDLSTM), and integrated convolutional neural network (CNN)-LSTM methods \cite{xie2016fully,shi2015end,he2015reading,sahu2015sequence}. In this paper, we investigate the most recently developed methods for online handwritten Chinese text recognition (OHCTR), which is an interesting research topic presenting the following challenges: a large character set, ambiguous segmentation, and variable-length input sequences.

Segmentation is the fundamental component of handwritten text recognition, and it has attracted the attention of numerous researchers \cite{seni1994external,kim1999architecture,cheriet2007character,wang2012approach,wang2012handwritten,zhou2013handwritten,zhou2014minimum}. Among the above-mentioned methods, over-segmentation \cite{wang2012approach,wang2012handwritten,zhou2013handwritten,zhou2014minimum}, i.e., an integrated segmentation-recognition method, is the most efficient method and still plays a crucial role in OHCTR. The basic concept underlying over-segmentation is to slice the input string into sequential character segments whose candidate classes can be used to construct the segmentation-recognition lattice \cite{wang2012handwritten}. Based on the lattice, path evaluation, which integrates the recognition scores, geometry information, and semantic context, is conducted to search for the optimal path and generate the recognition result. In practice, segmentation inevitably leads to mis-segmentation, which is barely rectifiable through post-processing and thus degrades the overall performance.

Segmentation-free methods are flexible alternative methods that completely avoid the segmentation procedure. HMMs and their hybrid variants \cite{bengio1999markovian,jiang2011novel,su2009off} have been widely used in handwritten text recognition. In general, the input string is converted into slices by sliding windows, followed by feature extraction and frame-wise prediction using an HMM. Finally, the Viterbi algorithm is applied to search for the best character string with maximum a posteriori probability.
However, HMMs are limited not only by the assumption that their observation depends only on the current state but also by their generative nature that generally leads to poor performance in labeling and classification tasks, as compared to discriminative models. Even though hybrid models that combine HMMs with other network architectures, including recurrent neural networks \cite{senior1998off,schenk2006novel} and multilayer perceptrons \cite{brakensiek1999performance,marukatat2001sentence}, have been proposed to alleviate the above-mentioned limitations by introducing context into HMMs, they still suffer from the drawbacks of HMMs.

The recent development of recurrent neural networks, especially LSTM \cite{graves2012supervised,graves2009novel,liwicki2007novel} and MDLSTM \cite{messina2015segmentation,graves2012supervised}, has provided a revolutionary segmentation-free perspective to the problem of handwritten text recognition. In general, LSTM is directly fed with a point-wise feature vector that consists of the $(x, y)$-coordinate and relative features, while it recurrently updates its hidden state and generates per-frame predictions for each time step. Then, it applies connectionist temporal classification (CTC) to perform transcription. 
It is worth noting that LSTM and MDLSTM have been successively applied to handwritten text recognition in Western languages, where the character set is relatively small (e.g., for English, there are only 52 classes; therefore it is easy to train the network). However, to the best of our knowledge, very few studies have attempted to address the problem of large-scale (where, e.g., the text lines may be represented by more than 2,500 basic classes of characters and sum up to more than 1 million character samples) handwritten text recognition problems such as OHCTR. 

Architectures that integrate CNN and LSTM exhibit excellent performance in terms of visual recognition and description \cite{pan2015jointly,venugopalan2015sequence}, scene text recognition \cite{shi2015end,he2015reading,sahu2015sequence}, and handwritten text recognition \cite{xie2016fully}. In text recognition problems, deep CNNs generate highly abstract feature sequences from input sequential data. LSTM is fed with such feature sequences  and generates corresponding character strings. Jointly training LSTM with CNN is straightforward and can improve the overall performance significantly. However, in the above-mentioned methods, the CNNs, specifically fully convolutional networks (FCNs), process the input string with only a fixed-size respective field in a sliding window manner, which we claim is inflexible for unconstrained written characters in OHCTR.

In this paper, we propose a novel solution (see Fig.~\ref{Fig:network}) that integrates path signature, a multi-spatial-context fully convolutional recurrent network (MC-FCRN), and an implicit language model to address the problem of unconstrained online handwritten text recognition. Path signature, a recent development in the field of the rough path theory \cite{lyonssystem,lyons2014rough,hambly2010uniqueness}, is a promising approach for translating variable-length pen-tip trajectories into offline signature feature maps in our system, because it effectively preserves the online information that characterizes the analytic and geometric properties of the path. Encouraged by recent advances in deep CNNs and LSTMs, we propose the MC-FCRN for robust recognition of signature feature maps. MC-FCRN leverages the multiple spatial contexts that correspond to multiple receptive fields in each time step to achieve strong robustness and high accuracy. Furthermore, we propose an implicit language model, which incorporate semantic context within the entire predicting feature sequence from both forward and reverse directions, to enhance the prediction for each time step.
The contributions of this paper can be summarized as follows:
\begin{itemize}
\item We develop a novel segmentation-free MC-FCRN to effectively capture the variable spatial contextual dynamics as well as the character information for high-performance recognition. Moreover, the main components of MC-FCRN, FCN, LSTM, and CTC can be jointly trained to benefit from one another, thereby enhancing the overall performance.
\item We propose an implicit language model that learns to model the output distribution given the entire predicting feature sequence.
Unlike the statistical language model that predicts the next word given only a few previous words, our implicit language model exploits the semantic context not only from the forward and reverse directions of the text but also with arbitrary text length.
\item Path signature, a novel mathematical feature set, bought from the rough path theory \cite{lyonssystem,lyons2014rough,hambly2010uniqueness} as a non-linear generalization of classical theory of controlled differential equations, is successfully applied to capture essential online information for long pen-tip trajectories using a sliding window-based method. Moreover, we investigate path signature for learning the variable online knowledge of the input string with different iterated integrals from both theoretical and empirical perspectives.
\end{itemize}

The remainder of this paper is organized as follows. Section~\ref{sec:related work} reviews the related studies.  Section~\ref{sec:path signature} formally introduces path signature and explains the sliding window-based method. Section~\ref{sec:Multi-spatial Context FCRN} details the network architecture of FCRN and its extended version, namely MC-FCRN. Section~\ref{sec:implicit_language_modeling} describes the proposed implicit language model and discusses the corresponding training strategy. Section~\ref{sec:Experimental} presents the experimental results. Finally, Section~\ref{sec:conclusion} concludes the paper.

\section{Related Work}
\label{sec:related work}
Feature extraction \cite{kimura1987modified,plamondon2000online,verma2004feature,bai2005study,biadsy2006online,liu2006online} plays a crucial role in traditional online handwritten text  recognition. The 8-directional feature \cite{kimura1987modified,bai2005study} is widely used in OHCTR owing to its excellent ability to express stroke directions. The projection of each trajectory point in eight directions is calculated in a 2-D manner and eight pattern images are generated accordingly. For further sophistication, Grave et al. \cite{graves2009novel} considered not only the $(x, y)$-coordinate and its relationship with its neighbors in the time series but also the spatial information from an offline perspective, thus obtaining 25 features for each point. However, the above-mentioned techniques have been developed empirically. Inspired by the theoretical work of Lyons and his colleagues \cite{hambly2010uniqueness,lyonssystem,lyons2014rough}, we applied path signature to translate the online pen-tip trajectories into offline signature feature maps that maintain the essential features for characterizing the online information of the trajectories. Furthermore, we can use truncated path signature in practical applications to achieve a trade-off between complexity and precision.

Yang et al. \cite{yang2015improved,yang2015dropsample} showed that the domain-specific information extracted by the aforementioned methods can improve the recognition performance with deep CNN (DCNN). However, DCNN-based networks are unable to handle input sequences of variable length in OHCTR. LSTM- and MDLSTM-based networks have an inherent advantage in dealing with such input sequences and demonstrate excellent performance in unconstrained handwritten text recognition \cite{graves2009novel,liwicki2007novel,bluche2016scan,bluche2016joint}. However, they are mainly focused on English text recognition, which involves an extremely small character set, including only letters and symbols. Our multi-spatial-context fully convolutional recurrent network (MC-FCRN) is based on recently developed deep learning methods that integrate LSTM and CNN in the field of visual captioning \cite{donahue2015long,venugopalan2015sequence} and scene text recognition \cite{he2015reading,shi2015end}. However, our MC-FCRN differs from these methods in that it uses multiple receptive fields of different scales to capture highly informative contextual features in each time step.
Such a multi-scale strategy originates from traditional methods. The pyramid match kernel \cite{grauman2005pyramid} maps features to multi-dimensional multi-resolution histograms that help to capture co-occurring features. The SIFT vectors \cite{lowe2004distinctive} search for stable features across all possible scales and construct a high-dimensional vector for the key points. Further, spatial pyramid pooling \cite{he2015spatial} allows images of varying size or scale to be fed during training and enhances the network performance significantly. GoogLeNet \cite{szegedy2015going} introduced the concept of ``inception'' whereby multi-scale convolution kernels are integrated to boost performance. We have drawn inspiration from these multi-scale methods to design our MC-FCRN. 

In general, language modeling is applied after feature extraction and recognition in order to improve the overall performance of the system \cite{zhou2014minimum,zhou2013handwritten,liu2006online,wang2012handwritten,wang2009integrating,wang2012approach}.
The recent development of neural networks, especially LSTM, in the field of language translation \cite{sutskever2014sequence} and visual captioning \cite{venugopalan2015sequence,pan2015jointly} has provided us with a new perspective of language models. 
To the best of our knowledge, neural networks were first applied to language modeling by Bengio et al. \cite{bengio2006neural}. Subsequently, Mikolov et al. \cite{mikolov2010recurrent} used recurrent neural network and Sundermeyer et al. \cite{sundermeyer2012lstm} used LSTM for language modeling. For language translation, Sutskever et al. \cite{sutskever2014sequence} used multilayered LSTM to encode the input text into a vector of fixed dimensionality and then applied another deep LSTM to decode the text in a different language. For visual captioning, Venugopalan et al. \cite{venugopalan2015sequence} and Pan et al. \cite{pan2015jointly} extracted deep visual CNN representations from image or video data and then used an LSTM as a sequence decoder to generate a description for the representations. 
Partially inspired by these methods, we developed our implicit language model to incorporate semantic context for recognition.
However, unlike the above-mentioned methods, which only derive context information from the past predicted text, our implicit language model learns to make predictions given the entire predicting feature sequence in both forward and reverse directions.

\begin{figure*}[th]
\label{Fig:signatureCompare}
\centering
\subfloat[]{
\label{FigureSigOri}
\includegraphics[width=0.48\textwidth]{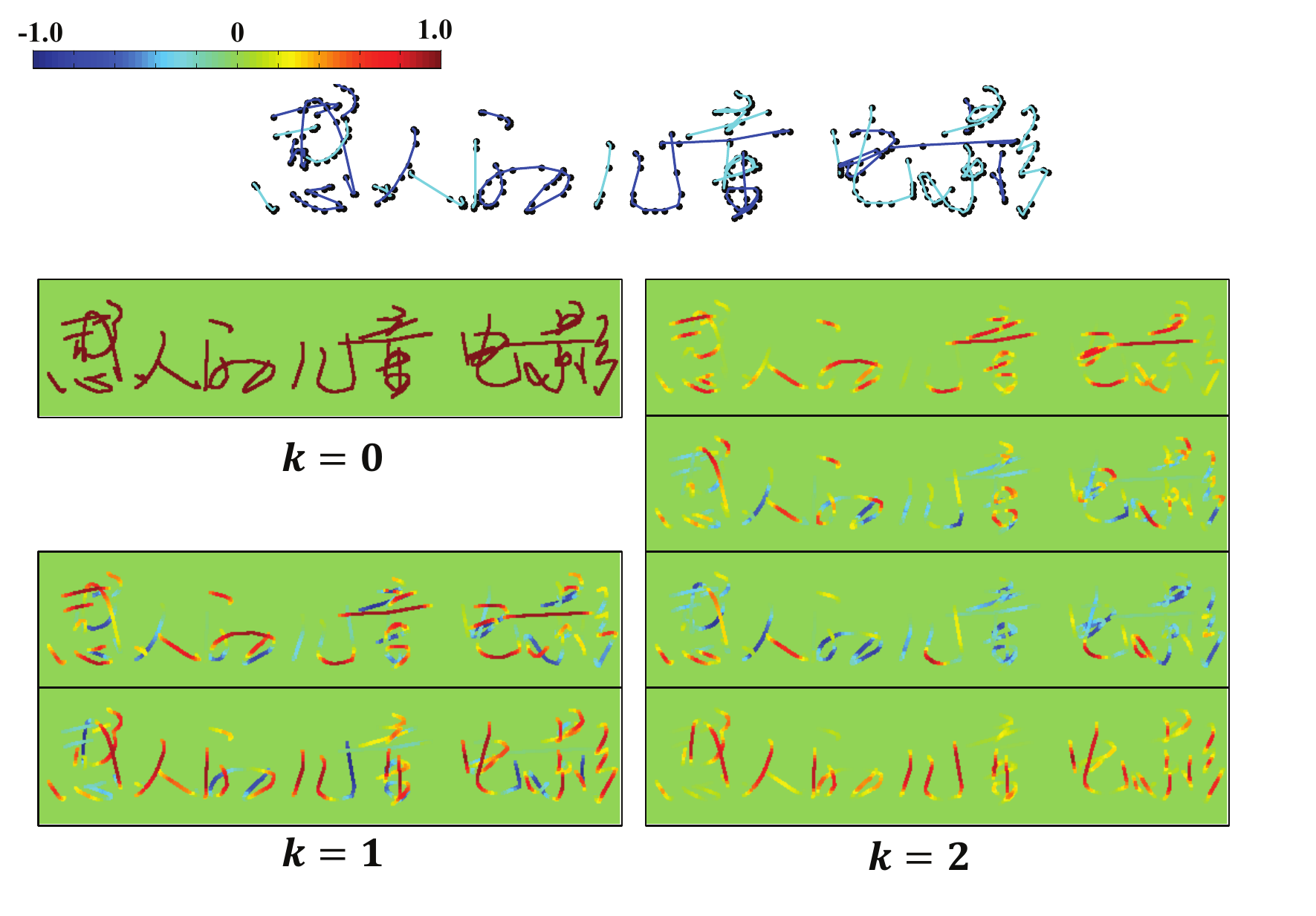}
}
\subfloat[]{
\label{FigureSigVir}
\includegraphics[width=0.48\textwidth]{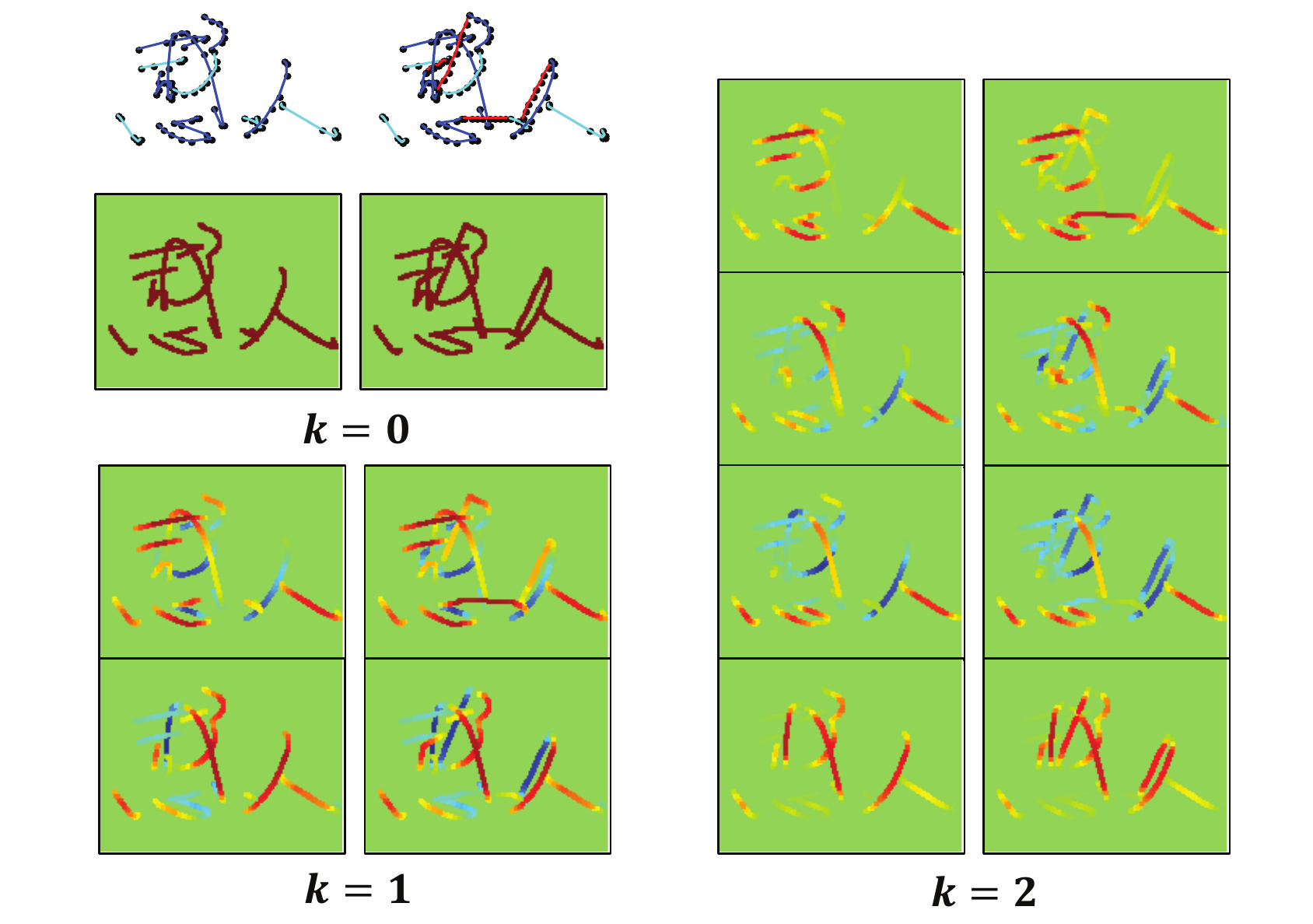}
}
\caption{(a) Path signature of the pen-tip trajectories. 
(b) Left: path signature of the original pen-tip trajectories; Right: path signature of the pen-tip trajectories with randomly added connections between adjacent strokes. It is notable that excepting for the additional connections, the original part of the sequential data has the same path signature (same color).}
\end{figure*}

\section{Path Signature}
\label{sec:path signature}
Proper translation of online data into offline feature maps while retaining most, or hopefully all, of the online knowledge within the pen-tip trajectory plays an essential role in online handwritten recognition.
Toward this end, we investigate path signature, which was pioneered by Chen \cite{chen1958integration} in the form of iterated integrals and developed by Lyons and his colleagues as a fundamental component of rough path theory \cite{hambly2010uniqueness,lyonssystem,lyons2014rough}.
We highlight the application of path signature in translating variable-length pen-tip trajectories into offline feature maps as a promising approach for retaining the online information for recognition. Path signature not only preserves the online dynamics but also captures many essential analytic and geometric properties of the path. Previous studies \cite{yang2015improved,yang2015dropsample} have adopted path signature for online feature extraction, but only at the character level. We go further by applying path signature to extremely long sequential data that usually consist of hundreds of thousands of points, using a sliding window-based method.  

Now, we briefly introduce path signature and discuss its application to OHCTR.
Consider the pen strokes of the online handwritten text collected from a writing plane $H \subset R^2$. Then, a pen stroke can be expressed as a continuous mapping denoted by $D:[a,b]\rightarrow H$ with $D=(D_{t}^1,D_{t}^2)$ and $t \in [a,b]$. For $k \ge 1$ and a collection of indexes $i_1,\cdots,i_k\in \{1,2\}$, the $k$-th fold \textsl{iterated integral} of $D$ along the index $i_1,\cdots,i_k$ can be defined by
\begin{equation}
\label{equ:signatureCal}
      P(D)_{a,b}^{i_1,\cdots,i_k} = \int_{a<t_1<\cdots <t_k<b} dD_{t_1}^{i_1},\cdots,dD_{t_k}^{i_k}.\\
\end{equation}
The signature of the path is a collection of all the iterated integrals of $D$:
\begin{align}
\label{equ:signatureCollect}
      P(D)_{a,b} = &(1,P(D)_{a,b}^1,P(D)_{a,b}^2,P(D)_{a,b}^{1,1},\nonumber\\
        &P(D)_{a,b}^{1,2},P(D)_{a,b}^{2,1},P(D)_{a,b}^{2,2},\cdots), 
\end{align}
where the superscripts of the terms $P(X)_{a,b}^{i_1,\cdots,i_k}$ run over the set of all \textsl{multi-indexes} 
\begin{equation}
  G=\{(i_1,...,i_k)|i_1,\cdots,i_k \in \{1,2\},k \ge 1\}.
\end{equation}
Then, the $k$-th iterated integral of the signature $P(D)_{a,b}^{(k)}$ is the finite collection of terms $P(D)_{a,b}^{i_1,\cdots,i_k}$ with multi-indexes of length $k$. 
More specifically, $P(D)_{a,b}^{(k)}$ is the $2^k$-dimensional vector defined by
\begin{equation}
  P(D)_{a,b}^{(k)}=(P(X)_{a,b}^{i_1,\cdots,i_k}|i_1,\cdots,i_k \in \{1,2\}).
\end{equation}
In \cite{hambly2010uniqueness}, it is proved that the whole signature of a path determines the path up to time re-parameterization. In practice we have to use the truncated signature feature, which can capture the global information on the path. Increasing the degree of truncated signature results in the exponential growth of dimension but may not always lead to significant marginal gain on the description of the path.

Next, we describe the practical calculation of path signature in OHCTR. 
For OHCTR, the pen-tip trajectories of the online handwritten text samples are represented by a sequence of sampling points. Since adjacent sampling points of text samples are connected by a straight line, $D=(D_{t}^1,D_{t}^2)$ and $t \in [t_1,t_2]$, the iterated integrals $P(D)_{t_1,t_2}^{(k)}$ can be calculated iteratively as follows:
\begin{eqnarray}
\label{equ:signatureCalDetail}
P(D)_{t_1,t_2}^{(k)}=\begin{cases} 1&,k = 0,\\
                   (P(D)_{t_1,t_2}^{(k-1)} \otimes \bigtriangleup_{t_1,t_2})/k &,k \ge 1,
          \end{cases}
\end{eqnarray}
where $\bigtriangleup_{t_1,t_2}:=D_{t_2}-D_{t_1}$ denotes the path displacement and $\otimes$ represents the tensor product.
Now, we define the \emph{concatenation} of two paths, $X: [t_1,t_2] \to H$ and $Y: [t_2,t_3] \to H$, in the writing plane $H$ as $X \ast Y : [t_1,t_3] \to H$, for which
\begin{eqnarray}
(X \ast Y)_t=\begin{cases} X_t&,t \in [t_1,t_2],\\
                   X_{t_2}+(Y_t-Y_{t_2})&,t \in [t_2,t_3].
          \end{cases}
\end{eqnarray}
\emph{Chen's identity} \cite{chen1958integration} formally expresses the calculation of path signature for the concatenation of two paths as
\begin{equation}
  \label{equ:sig_chen_identity}
  P(X \ast Y)_{t_1,t_3}=P(X)_{t_1,t_2} \otimes P(Y)_{t_2,t_3}.
\end{equation}
Based on Eq.~\eqref{equ:signatureCalDetail} and Eq.~\eqref{equ:sig_chen_identity}, the path signature of pen-tip trajectories of arbitrary length can be calculated.
Fig.~\ref{FigureSigOri} shows the $0, 1, 2$-th iterated integral signature feature maps of the pen-tip trajectories to better illustrate the concept of path signature.

In this paper, we adopt a sliding window-based method to extract signature features that robustly represent online dynamics for text recognition. Many previous studies attempted to directly extract path signature at the stroke level, because stroke is the basic path element in OHCTR. However, online sequential data are cursively written and generally have additional connections between adjacent strokes within a character or between characters. 
Such additional connections may degrade recognition performance because they generally lead to mutable path signature features.
Therefore, in this paper, we limit the path to a small window of points and extract path signature using these windows paths.

Given a pen stroke $D$, we obtain a zone of interest using the following sliding window function:
\begin{eqnarray}
g(w)=\begin{cases} 1,&\frac{-W}{2}<w<\frac{W}{2},\\
                   0,&\text{otherwise},
          \end{cases}
\end{eqnarray}
where $W$ represents the window width. In our experiment, we set $W=9$, because it offers robustness against local distortion while being sufficiently sensitive to capture the local geometric properties of the path.
Let $V$ denote the shift step of the window. Then, the window path in step $i$ can be represented by $D\cdot g(w-i \cdot V)$.
Thus, we can use Eqs.~\eqref{equ:signatureCalDetail} and \eqref{equ:sig_chen_identity} to extract the path signature of the window path.
Note that the shift step $V$ determines whether there is overlap between adjacent windows. In our experiments, we set $V$ to a single point and used the path signature of the window path to represent the midpoint feature of the window path. 
In this manner, we can extract the online information from pen-tip trajectories with strong local invariance and robustness. 
Fig.~\ref{FigureSigVir} shows that, although connections are randomly added between adjacent strokes within a character or between characters, the impact on the path signature of the original input string is not significant, which proves that path signature based on windows has excellent local invariance and robustness.

\section{Multi-Spatial-Context FCRN} 
\label{sec:Multi-spatial Context FCRN}

\begin{figure}[t]
\centering
\includegraphics[width=0.5\textwidth]{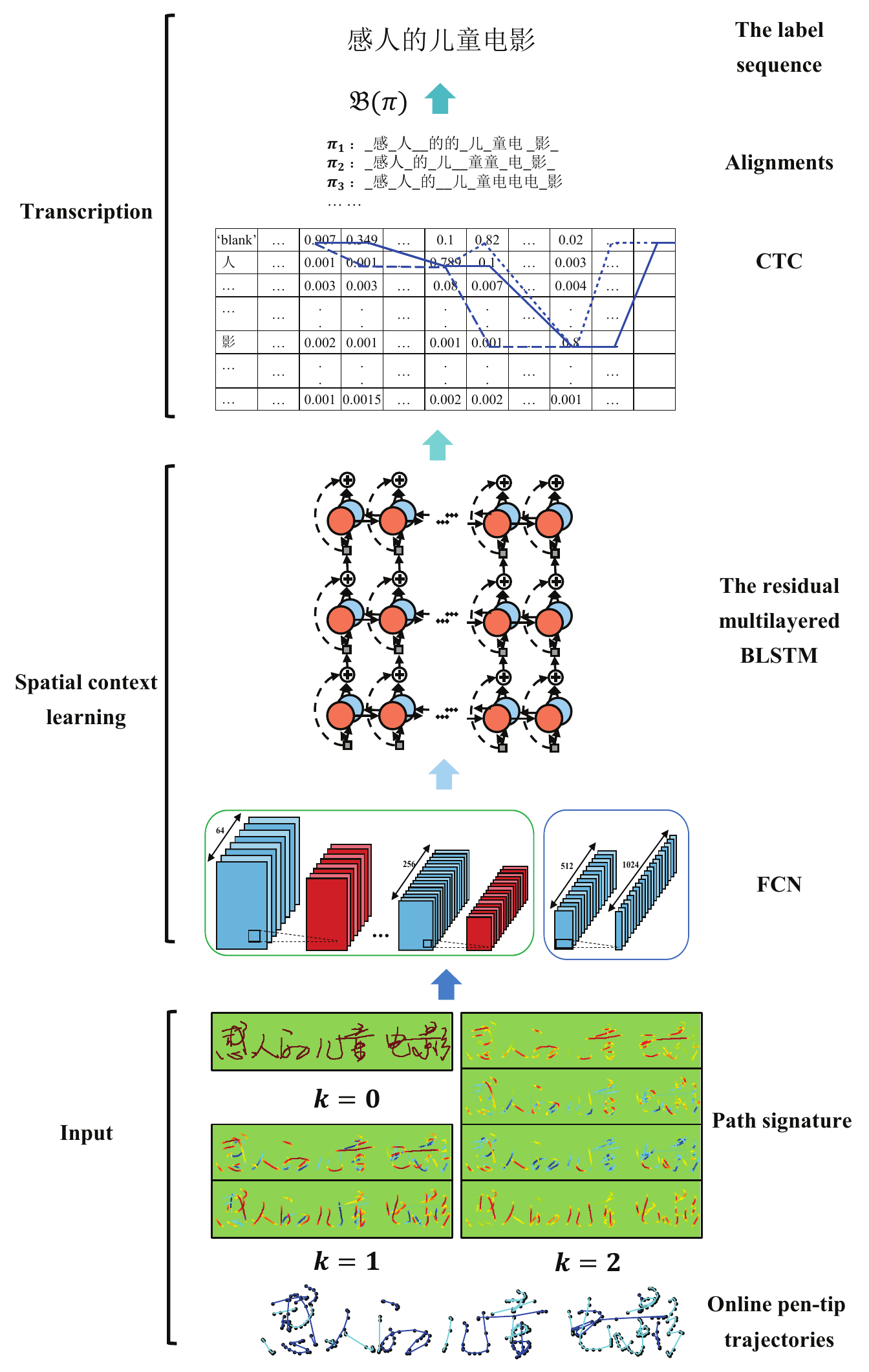}
\caption{Illustration of FCRN. Given the path signature feature maps of the online pen-tip trajectories (the order of the characters is actually shuffled during the training procedure), a fully convolutional network (FCN) produces a deep feature sequence in which each frame represents the feature vector of a receptive field on the
signature feature maps. Then, a residual multilayered BLSTM generates a prediction distribution for each frame in the feature sequence. Finally, the transcription layer derives a label sequence from the per-frame predictions for recognition.}
\label{Fig:total_framework}
\end{figure}

Unlike character recognition, where it is easy to normalize characters to a fixed size, text recognition is complicated because it involves input sequences of variable length, such as feature maps and online pen-tip trajectories. 
We propose a new fully convolutional recurrent network (FCRN) for spatial context learning to overcome this problem by leveraging a fully convolutional network, a recurrent neural network, and connectionist temporal classification, all of which naturally take inputs of arbitrary size or length (see Fig.~\ref{Fig:total_framework}).
Furthermore, we extend our FCRN to multi-spatial-context FCRN (MC-FCRN), as shown in Fig.~\ref{Fig:network}, to learn multi-spatial-context knowledge from complicated signature feature maps.
In the following subsections, we briefly introduce the basic components of FCRN and explain their roles in the architecture. Then, we demonstrate how MC-FCRN performs multi-spatial-context learning for the OHCTR problem.

\subsection{Fully Convolutional Recurrent Network}
\subsubsection{Fully Convolutional Network}
\label{subsubsec:fully_convolutional_network}
DCNNs exhibit excellent performance in computer vision applications such as image classification \cite{he2015deep,he2015spatial}, scene text recognition \cite{shi2015end,he2015reading}, and visual description \cite{venugopalan2015sequence,pan2015jointly}. 
Following the approach of Long et al. \cite{long2015fully}, we remove the original last fully connected classification layer from DCNNs to construct a fully convolutional network. Fully convolutional networks not only inherit  the ability of DCNNs to learn powerful and interpretable image features but also adapt to variable input image size and generate corresponding-size feature maps. It is worth noting that such CNN feature maps contain strong spatial order information from the overlap regions (known as receptive fields) of the original feature maps. Such spatial order information is very important and can be leveraged to learn spatial context to enhance the overall performance of the system.
Furthermore, unlike image cropping or sliding window-based approaches, FCNs eliminate redundant computations by sharing convolutional response maps layer by layer to achieve efficient inference and backpropagation.

\subsubsection{Multilayered BLSTM}
Recurrent neural networks (RNNs), which are well known for the self-connected hidden layer that recurrently transfers information from output to input, have been widely adopted to learn continuous sequential features. Recently, long short-term memory (LSTM)\cite{hochreiter1997long}, a variant of RNN that overcomes the gradient vanishing and exploding problem, has demonstrated excellent performance in terms of learning complex and long-term temporal dynamics in applications such as language translation\cite{bahdanau2014neural}, visual description \cite{venugopalan2015sequence,pan2015jointly}, and text recognition \cite{he2015reading,shi2015end}. 
Owing to the inherent sequential nature of the input string data, LSTM is particularly well suited for context information learning in OHCTR.
Given the deep FCN feature sequence, our LSTM recursively takes one frame from the input feature sequence, updates the hidden states, and predicts a distribution for further transcription.
We emphasize the following advantages of LSTM in the OHCTR problem. 
First, LSTM is not limited to fixed-length inputs or outputs, allowing for modeling of sequential data of arbitrary length.
Furthermore, LSTM naturally captures the contextual information from a sequence \cite{shi2015end}, making the text recognition process more efficient and reliable than processing each character independently.
Finally, LSTM can be jointly trained with FCN in a unified network to improve the overall text recognition performance.

Standard LSTM only uses past contextual information in one direction, which is inadequate for the OHCTR problem, where bidirectional contextual knowledge is accessible.
Bidirectional LSTM (BLSTM) facilitates the learning of complex context dynamics in both forward and reverse directions, thereby outperforming unidirectional networks significantly.
As suggested by Pascanu et al.\cite{pascanu2013construct}, we stacked multiple BLSTMs to access higher-level abstract information in temporal dimensions for further transcription.

\subsubsection{Transcription} 
Connectionist temporal classification (CTC), which facilitates the use of FCN and LSTM for sequential training without requiring any prior alignment between input images and their corresponding label sequences, is adopted as the transcription layer in our framework.
Let $C$ represent all the characters used in this problem and let ``blank'' represent the null emission. Then, the character set can be denoted as $C'=C \cup \{blank\}$.
Given input sequences $\bm{u}= (u_1,u_2,\cdots,u_T)$ of length $T$, where $u_t \in R^{|C'|}$, we can obtain an exponentially large number of label sequences of length $T$, refered to as alignments $\bm{\pi}$, by assigning each time step a label and concatenating the labels to form a label sequence.
The probability of alignments is given by
\begin{equation}
  \label{Equ:alignmentCal}
    p(\bm{\pi}|\bm{u}) = \prod_{t=1}^T p(\pi_t,t|\bm{u}).
\end{equation}
Alignments can be mapped onto a transcription $\bm{l}$ by applying a sequence-to-sequence operation $\mathscr{B}$, which first removes the repeated labels and then removes the blanks. For example, ``tree'' can be obtained by $\mathscr{B}$ from ``\_tt\_r\_ee\_e'' or ``\_t\_rr\_e\_eee\_''. The total probability of a transcription can be calculated by summing the probabilities of all alignments corresponding to it:
\begin{equation}
\label{Equ:transcriptionCal}
    p(\bm{l}|\bm{u}) = \sum_{\bm{\pi}:\mathscr{B}(\bm{\pi})=\bm{l}} p(\bm{\pi}|\bm{u}).
\end{equation}

As suggested by Graves and Jaitly\cite{graves2014towards}, since the exact position of the labels within a particular transcription cannot be determined, we consider all the locations where they could occur, thereby allowing a network to be trained via CTC without pre-segmented data.
A detailed forward-backward algorithm to efficiently calculate the probability in Eq.~\eqref{Equ:transcriptionCal} was proposed by Graves \cite{graves2012supervised}.

\subsection{Learning Multiple Spatial Context with Multi-Spatial-Context FCRN (MC-FCRN)}
FCNs generate deep corresponding-size feature maps that are closely related to the input images (e.g., signature feature map). 
We simplified our problem such that the FCN is only required to output a corresponding-length feature sequence of the original signature feature maps by collapsing across the vertical dimension of the top-most feature maps. 
Thus, the output $N$ feature maps of the FCN, which has a height of one pixel and variable width $W$, can easily be transformed into the feature sequence with length $T=W$ and dimension $N$, in the sense that the $t$-th time step of the feature sequence involves the concatenation of all the $t$-th pixel values of the output FCN feature maps. 
Such a feature sequence possesses strong ordered spatial information, because successive frames in the sequence correspond to the overlapped receptive fields of the input images.

\subsubsection{Spatial Context}

\begin{figure*}[t]
\centering
\includegraphics[width=0.9\textwidth]{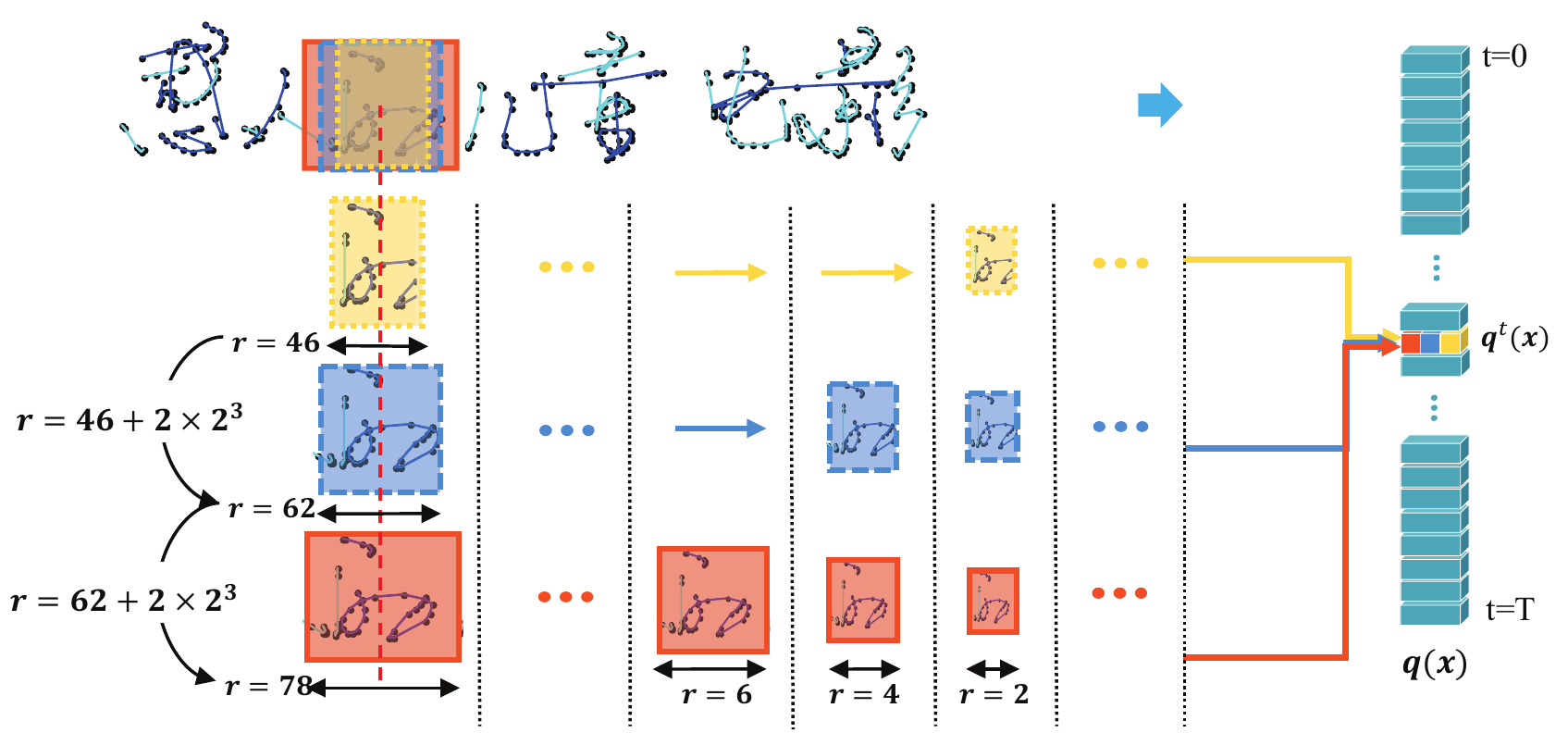}
\caption{Illustration of multiple spatial contexts. Different receptive fields in the same time step have the same center position, and their region sizes should satisfy Eq.~\eqref{equ:multi-spatial_mapping}.}
\label{Fig:multi-spatial-network}
\end{figure*}

First, we introduce a simple model having three components: FCN (denoted by $f$ with parameters $\bm{\omega}_f$), fully connected layers (denoted by $g$ with parameters $\bm{\omega}_g$), and CTC.
Given signature feature maps $\bm{x}=(x_1, x_2, \cdots, x_T)$, where $x_t$ represents the receptive field of the $t$-th time step, the objective of this model is to learn to make a prediction for each receptive field:
\begin{align}
\label{equ:spatial_FCN}
&o(z_t,x_t) = g(z_t,f(x_t))\\
&s.t.\quad
\begin{cases}
\sum^{|C'|}_{i=1} g(z_t=C'_i,f(x_t)) = 1, \nonumber\\
g(z_t,f(x_t)) > 0.
\end{cases}
\end{align}
where $z_t$ represents the prediction of the $t$-th time step and $o(z_t,x_t)$ models the probability distribution over all the words in $C'$ given the receptive field $x_t$ in the $t$-th time step.
Since each frame in the FCN feature sequence represents a distribution over the character set without considering any other feature vector, this simple model can hardly incorporate any spatial context for recognition.
LSTM inherently possesses the advantage of processing sequential data; thus, we consider it as a good choice for capturing the spatial context within the FCN feature sequence.
In this work, we jointly apply BLSTM (denoted by $h$ with parameters $\bm{\omega}_h$) and residual connection \cite{he2015deep} to the output of FCN in order to learn spatial context from the FCN feature sequence. 
In practice, multiple BLSTMs and residual connections are stacked to construct a residual multilayered BLSTM for deeper spatial context learning. Denoting the $l$-th BLSTM layer output as $q_l(\bm{x})$, we have
\begin{equation}
\label{Equ:LSTM_residual}
  q_{l}(\bm{x}) = h(q_{l-1}(\bm{x}))+q_{l-1}(\bm{x}).
\end{equation}
Iteratively applying $q_{l}(\bm{x}) = h(q_{l-1}(\bm{x}))+q_{l-1}(\bm{x})= h(q_{l-1}(\bm{x})) + h(q_{l-2}(\bm{x})) + q_{l-2}(\bm{x})$ to Eq.~\eqref{Equ:LSTM_residual}, we get
\begin{equation}
\label{Equ:LSTM_all}
    q_{L}(\bm{x}) =  q_0(\bm{x}) + \sum^{L-1}_{i=1} {h(q_i(\bm{x}))},
\end{equation}
where $L$ is the total number of layers of the residual multilayered BLSTM. Note that $q_0(\bm{x}) =(f(x_1),f(x_2) \dots f(x_T))$ is the FCN output. 
Thus, we have actually constructed our FCRN.
The objective of the FCRN is to learn to make a prediction for each time step given the entire input signature feature maps:
\begin{align}
\label{equ:spatial_FCN_LSTM}
&o(z_t,\bm{x}) = g(z_t,q^{t}_L(\bm{x}))\\
&s.t.\quad
\begin{cases}
\sum^{|C'|}_{i=1} g(z_t=C'_i,q^{t}_L(\bm{x})) = 1, \nonumber\\
g(z_t,q_L) > 0.
\end{cases}
\end{align}
where the overall system has parameters $\bm{\omega} = (\bm{\omega}_f, \bm{\omega}_h, \bm{\omega}_g)$, 
$q^{t}_L(\bm{x})$ is the $t$-th time step of the output feature sequence of the residual multilayered BLSTM,
and $o(z_t,\bm{x})$ models the probability distribution over all the words in $C'$ in the $t$-th time step given the entire input signature feature maps $\bm{x}$.
Comparing Eq.~\eqref{equ:spatial_FCN} with Eq.~\eqref{equ:spatial_FCN_LSTM}, we can see that the simple model makes a prediction for each time step given only the $t$-th receptive field. 
To the contrary, FCRN focuses on the entire signature feature maps; thus, it is able to incorporate spatial context to enhance recognition performance.

Now, suppose $p(\pi_t,t|\bm{x}) = g(z_t=\pi_t,q^{t}_L(\bm{x}))$. By Eq.~\eqref{Equ:alignmentCal}, the probability of alignments can be represented by:
\begin{equation}
  \label{Equ:alignmentCalWithFCRN}
    p(\bm{\pi}|\bm{x}) = \prod_{t=1}^T g(z_t=\pi_t,q^{t}_L(\bm{x})).
\end{equation}
Then, the total probability of a transcription can be calculated by applying Eq.~\eqref{Equ:alignmentCalWithFCRN} to Eq.~\eqref{Equ:transcriptionCal}:
\begin{equation}
\label{Equ:transcriptionCalWithFCRN}
    p(\bm{l}|\bm{x}) = \sum_{\bm{\pi}:\mathscr{B}(\bm{\pi})=\bm{l}} \prod_{t=1}^T g(z_t=\pi_t,q^{t}_L(\bm{x})).
\end{equation}
The training is achieved by searching for $\bm{\omega}$ that minimizes the negative penalized log-likelihood:
\begin{equation}
    L(Q) = -\sum_{(\bm{x},\bm{l}) \subset Q} ln\{\sum_{\bm{\pi}:\mathscr{B}(\bm{\pi})=\bm{l}} \prod_{t=1}^T g(z_t=\pi_t,q^{t}_L(\bm{x});\bm{\omega})\}+R(\bm{\omega})
\end{equation}
where $\bm{l}$ is the label sequence, $Q$ represents the training set, and $R(\bm{\omega})$ denotes the regularization term. In our experiment, $R$ is a weight decay penalty implemented with $L2$ regularization.

\subsubsection{Receptive Field and Its Role with Multi-Spatial Context}

A receptive field is a rectangular local region of input images that can be properly represented by a highly abstract feature vector in the output feature sequence of FCN. 
Let $r_l$ represent the local region size (width/height) of the $l$-th layer, and let the $(x_l,y_l)$-coordinate denote the center position of this local region. Then, the relationship of $r_l$ and $(x_l,y_l)$-coordinate between adjacent layers can be formulated as follows:
\begin{equation}
\label{Equ:FCN_detail}
    \begin{split}
        r_l&=(r_{l+1}-1)\times m_l+k_l,\\
      x_l&=m_l \times x_{l+1}+(\frac{k_l-1}{2}-p_l),\\
      y_l&=m_l \times y_{l+1}+(\frac{k_l-1}{2}-p_l),       
    \end{split}
\end{equation}
where $k$ is the kernel size, $m$ is the stride size, and $p$ is the padding size of a particular layer. Recursively applying Eq.~\eqref{Equ:FCN_detail} to adjacent layers in the FCN from the last response maps down to the original image should yield the region size and the center coordinate of the receptive field that corresponds to the related feature vector of the FCN feature sequence.

In the following, we explain how to generate receptive fields with different scales for each time step.
We observe that the size of the receptive field is sensitive to the kernel size. 
Assume that the kernel size is increased from $k_l$ to $k_l+\Delta k_l$. We can derive the following mapping from Eq.~\eqref{Equ:FCN_detail}: $r_{l-1}=r'_{l-1}+\Delta k_l \times m_{l-1}$, where $r'_{l-1}$is the original region size of the $(l-1)$-th layer. 
Thus, we have 
\begin{equation}
\label{equ:multi-spatial_mapping}
  r_0=r'_0+\Delta k_l \times \prod^{l-1}_{i=1} m_i. 
\end{equation}
In other words, if we increase the kernel size of the $l$-th layer by $\Delta k_l$, then the receptive field will be enlarged by $\Delta k_l \times \prod^{l-1}_{i=1} m_i$.
As shown in Fig.~\ref{Fig:multi-spatial-network}, our MC-FCRN takes advantage of such a phenomenon and fuses receptive fields of different scales for each time step.
Further, note that when $k_l=2p_l+1$ and $m_l=1$, the center position (i.e., $(x_l,y_l)$-coordinate) of the receptive field does not change from higher layers to lower layers. Therefore, receptive fields with different scales in the same time step have the same center position, which ensures that multiple spatial contexts are incorporated while confusion is avoided.

Let $\bm{q}(\bm{x})$ represent the concatenation of the output feature sequences of the residual multilayered BLSTMs with receptive fields of different scales (see Fig.~\ref{Fig:multi-spatial-network}).
Formally, the objective of our MC-FCRN is to learn to make a prediction for each time step given the entire input signature feature maps:
\begin{align}
\label{equ:mulit-spatial_FCN_LSTM}
&o(z_t,\bm{x}) = g(z_t,\bm{q}^{t}(\bm{x}))\\
&s.t.\quad
\begin{cases}
\sum^{|C'|}_{i=1} g(z_t=C'_i,\bm{q}^{t}(\bm{x})) = 1, \nonumber\\
g(z_t,\bm{q}^{t}(\bm{x})) > 0.
\end{cases}
\end{align}
where $\bm{q}^t(\bm{x}) $ is the concatenation of $t$-th time step of the output feature sequence of the residual multilayered BLSTM with receptive fields of different scales, and $o(z_t,\bm{x})$ models the probability distribution over all the words in $C'$ in the $t$-th time step given the entire input signature feature maps $\bm{x}$.



\section{Implicit Language Modeling}
\label{sec:implicit_language_modeling}

We say that a system is an implicit language model if it does not directly learn the conditional probabilities of the next word given previous words, but implicitly incorporates lexical constraints and prior knowledge about the language to improve the system performance. The network architecture of our implicit language model
consists of three components: the embedding layer, the language modeling layer, and the transcription layer.
Given the predicting feature sequence $\bm{s}=(s_1, s_2, \cdots, s_T)$ from the multi-spatial-context FCRN, the objective of the implicit language model is to learn to make a prediction for each time step given the entire input sequence:
\begin{align}
\label{equ:implicit_LM}
&f(z_t,\bm{s}) = U(z_t,(M(s_1), M(s_2), \cdots, M(s_T))) \\
&s.t.\quad
\begin{cases}
\sum^{|C'|}_{i=1} f(z_t=C'_i|\bm{s}) = 1, \nonumber\\
f(z_t|\bm{s}) > 0.
\end{cases}
\end{align}
where $f(z_t,\bm{s})$ models the probability distribution over all the words in $C'$ in the $t$-th time step given the entire predicting feature sequence $\bm{s}$, while the mapping $M$ and the probability function $U$ represent two successive processing stages that constitute the prediction procedure of the implicit language model.
In the first stage, the mapping $M$, implemented by the embedding layer, translates the input sequence $\bm{s}=(s_1, s_2, \cdots, s_T)$ into real vectors $(M(s_1), M(s_2), \cdots, M(s_T))$, where $M(s_t) \in R^m$. 
Note that the mapping $M$ differs from the mapping $C$ \cite{bengio2006neural} in traditional neural language models, because the embedding mapping used here takes the input of a predicting feature vector, not just a one-hot vector. 
In the second stage, the probability function $U$, maps the embedding vectors for words in context $(M(s_1), M(s_2), \cdots, M(s_T))$ to a conditional probability distribution over all the words in $C'$, i.e., the $i$-th element of the output vector of $U$ estimates the probability $p(z_t=C'_i|(M(s_1), M(s_2), \cdots, M(s_T)))$.
Then, we can represent the function $f(z_t,\bm{s})$ by the composition of mappings $M$ and $U$ as follows: 
\begin{equation}  
  f(z_t = C'_i,\bm{s}) = U(C'_i,(M(s_1), M(s_2), \cdots, M(s_T)))
\end{equation}
In this paper, we implemented the embedding layer as a fully connected layer that can be represented by a $|C'| \times m$ matrix $\bm{\theta}_M$. The function $U$ with parameters $\bm{\theta}_U$ is implemented by multilayered BLSTM for learning long-term context from the forward and reverse directions. The overall parameters for the implicit language model $f$ are given by $\bm{\theta} = (\bm{\theta}_M, \bm{\theta}_U)$.

Now, suppose $p(\pi_t,t|\bm{s}) = f(z_t=\pi_t|\bm{s})$. By Eq.~\eqref{Equ:alignmentCal}, the probability of alignments can be represented by:
\begin{equation}
  \label{Equ:alignmentCalWithMultiFCRN}
    p(\bm{\pi}|\bm{s}) = \prod_{t=1}^T f(z_t=\pi_t|\bm{s}).
\end{equation}
Then, the total probability of a transcription can be calculated by applying Eq.~\eqref{Equ:alignmentCalWithMultiFCRN} to Eq.~\eqref{Equ:transcriptionCal}:
\begin{equation}
\label{Equ:transcriptionCalWithMultiFCRN}
    p(\bm{l}|\bm{s}) = \sum_{\bm{\pi}:\mathscr{B}(\bm{\pi})=\bm{l}} \prod_{t=1}^T f(z_t=\pi_t|\bm{s}).
\end{equation}
The training is achieved by searching for $\bm{\theta}$ that minimizes the negative penalized log-likelihood:
\begin{equation}
\label{Equ:LossWithImplicitLM}
    L(Q) = -\sum_{(\bm{s},\bm{l}) \subset Q} ln\{\sum_{\bm{\pi}:\mathscr{B}(\bm{\pi})=\bm{l}} \prod_{t=1}^T f(z_t=\pi_t|\bm{s};\bm{\theta})\}+R(\bm{\theta})
\end{equation}
where $R(\bm{\theta})$ is the regularization term.

\subsection{Interesting Properties of Implicit Language Model}
\begin{itemize}
\item Our implicit language model offers the unique advantage of leveraging semantic context from both directions of the text, significantly outperforming language models that only predict the conditional probability of the next word given previous words in one direction (see Section~\ref{sec:Effect of semantic context}). Furthermore, because LSTM can capture long-term complicated dynamics in the sequence, our implicit language model has the potential to learn semantic context from the entire sequence to enhance recognition.
\item The predicting feature sequence contains information that indicates not only the predicted labels but also the confidence about their prediction, providing much more information than a simple one-hot vector \cite{bengio2006neural}. Thus, the implicit language model is able to improve network performance by exploiting the confidence information of the predicted labels in addition to their semantic context knowledge.
\end{itemize}

\subsection{Training Strategy}
\label{sec:training_strategy}
Given a training instance $(\bm{x},\bm{l})$, we feed the trained MC-FCRN with the signature feature maps $\bm{x}$ to obtain a predicting feature sequence of length $T$. Then, the feature sequence with label $\bm{l}$ is used to train the implicit language model with parameter $\bm{\theta}$.
The training strategy is summarized in Algorithm~\ref{AlgorithmTraining Strategy}.
Note that the training set for the implicit language model should contain semantic information, i.e., the characters should be understandable in context. In fact, the original training set of CASIA-OLHWDB \cite{liu2011casia} contains semantic information, but it is not sufficiently large. Therefore, we expand our training set for the implicit language model using the sample synthesis technique. 
However, our training set for MC-FCRN training does not contain any semantic knowledge. Actually, we shuffle the order of the characters in each training instance to achieve this effect.
There are two main reasons for using different training sets during the training procedure of MC-FCRN and the implicit language model. 
First, our MC-FCRN and implicit language model can concentrate on learning spatial context and semantic context, respectively. 
If we directly learn spatial-semantic context in a unified network, then such a network may heavily overfit the context information of the training set.
Second, because the training set of CASIA2.0-2.2 has the same corpus as the test set, we cannot use the samples from the training set directly for training, as it may lead to unfair comparison with the results of other methods.

\begin{algorithm}[t]
\caption{Training strategy for implicit language model}
\label{AlgorithmTraining Strategy}
\begin{algorithmic}[1]
\REQUIRE ~~\\
  Iteration number $t = 0$;\\
  Training set $Q={(\bm{x},\bm{l})_1,\cdots,(\bm{x},\bm{l})_N}$;\\
  A trained MC-FCRN;
\ENSURE  ~~\\
  Network parameters of implicit language model $\bm{\theta}$
\REPEAT
  \STATE $t \gets t+1$.
  \STATE Randomly select a subset of samples from training set $Q$.
  \FOR {each training sample $(\bm{x},\bm{l})_i$}
  \STATE Do forward propagation with MC-FCRN to calculate the object function Eq.~\eqref{equ:mulit-spatial_FCN_LSTM};
  \STATE Take output of MC-FCRN as input of implicit language model;
  \STATE Do forward propagation with implicit language model and calculate the object function Eq.~\eqref{Equ:LossWithImplicitLM}.
  \ENDFOR
  \STATE $\Delta \bm{\theta} = 0$.
  \FOR {each training sample $(\bm{x},\bm{l})_i$}
  \STATE Run backpropagation through time to obtain gradient $\Delta \bm{\theta}_i$ with respect to the network parameters;
  \STATE Accumulate gradients: $\Delta \bm{\theta} := \Delta \bm{\theta} + \Delta \bm{\theta}_i$.
  \ENDFOR 
  \STATE $\bm{\theta}^t := \bm{\theta}^{t-1} - \eta \Delta \bm{\theta}$.
\UNTIL{convergence}
\end{algorithmic}
\end{algorithm}

\subsection{Statistical Language Model} 
\label{sub:explicit_language_model}
In the post-processing procedure, the language model plays a significant role in decoding the prediction
sequence \cite{zhou2014minimum,zhou2013handwritten,wang2012handwritten}. Beside the proposed implicit language model, we also use the traditional statistical language model (a trigram language model) in our experiment to jointly work with the implicit language model to further improve the overall recognition performance.

\section{Experiments}
\label{sec:Experimental}
To evaluate the effectiveness of the proposed system, we conducted experiments on the standard benchmark dataset CASIA-OLHWDB \cite{liu2011casia} and the ICDAR2013 Chinese handwriting recognition competition dataset \cite{yin2013icdar} for unconstrained online handwritten Chinese text recognition.

\subsection{Databases}
\label{sec:databases}
In the following experiments, we used the training set of CASIA-OLHWDB \cite{liu2011casia}, including both unconstrained text lines and isolated characters, as our training data.
The training set of CASIA2.0-2.2 (a subset of CASIA-OLHWDB for OHCTR problem) contains 4072 pages of handwritten texts, with 41,710 text lines, including 1,082,220 characters of 2650 classes.
We randomly split the training set into two groups, with approximately 90\% for training and the remainder for validation and further parameter learning for language modeling.
Two popular benchmark datasets for unconstrained online handwritten Chinese text recognition were used for performance evaluation, i.e., the test set of CASIA2.0-2.2 (Dataset-CASIA) and the test set of the online handwritten Chinese text recognition task of the ICDAR 2013 Chinese handwriting recognition competition\cite{yin2013icdar} (Dataset-ICDAR). Dataset-CASIA contains 1020 text pages, including 268,924 characters of 2626 classes, while Dataset-ICDAR contains 3432 text lines, including 91,576 characters of 1375 classes. 
It is worth noting that Dataset-ICDAR is smaller than the reported one by around 2.02\%, as outlier characters not covered by the training data were removed.

For language modeling, we conducted experiments using both the implicit language model and the statistical language model. Three corpora were used in this paper (see Table~\ref{TableCorpora}): the PFR corpus\cite{PFR}, which contains news text from the 1998 People's Daily corpus;
the PH\cite{PH} corpus, which contains news text from the People's Republic of China's Xinhua news recorded between January 1990 and March 1991;
and the SLD corpus\cite{SG}, which contains news text from the 2006 Sogou Labs data.
Because the total amount of the Sogou Labs data was extremely large, we only used an extract in our experiments.
For statistical language modeling, we used the SRILM toolkit\cite{stolcke2002srilm} to build our language model.
\begin{table}[h]
\caption{Character Information in the Corpora}
\label{TableCorpora}
\centering
\begin{tabular}{c|cc}
\hline
corpora&\#characters&\#class\\
\hline
PFR&2,199,492 &4,689\\
PH&3,697,028&4,722\\
SLD&56,279,692&6,882\\
\hline
\end{tabular}
\end{table}
  
\subsection{Experimental Setting}
\label{sec:experimental_setting}

The detailed architecture of our MC-FCRN and implicit language model is shown in Fig.~\ref{Fig:networkArchicture}. 
Batch normalization \cite{ioffe2015batch} was applied after all but the first two convolutional layers in order to achieve faster convergence and avoid over-fitting.
To accelerate the training process, our network was trained with shorter texts segmented from text lines in the training data, which could be normalized to the same height of 126 pixels while retaining the width at fewer than 576 pixels.
In the test phase, we maintained the same height but increased the width to 2400 pixels in order to include the text lines from the test set.
As discussed in Section~\ref{sec:training_strategy}, we used the data synthesis technique based on the isolated characters of CASIA-OLHWDB and the corpora listed in Table~\ref{TableCorpora} in order to enrich the training data. Note that, during the MC-FCRN training, we randomly shuffled the order of characters in each text training sample to discard the semantic context for fairness.
We constructed our experiments within the CAFFE\cite{jia2014caffe} deep learning framework, in which LSTM is implemented following the approach of Venugopalan et al.\cite{venugopalan2015sequence} while the other processes are contributed by ourselves.
Further, we used AdaDelta as the optimization algorithm with $\rho=0.9$. The experiments were conducted using GeForce Titan-X GPUs.
For performance evaluation, we used the correct rate (CR) and accuracy rate (AR) as performance indicators, as specified in the ICDAR 2013 Chinese handwriting recognition competition\cite{liuicdar}.

\subsection{Experimental Results}
\label{sec:Experimental_results}
\subsubsection{Effect of Path Signature}
\label{sec:Effect of iterated integral of path signature}

\begin{table}[b]\scriptsize
\caption{Effect of Path Signature (Percent)}
\label{Table:the effect of iterated integral of path signature}
\centering
\begin{threeparttable}
\begin{tabular}{cIccIcccc}
\whline
Path signatures&CR&AR&\textsl{Chinese}&\textsl{Symbol}&\textsl{Digit}&\textsl{Letter}\\
\whline
Sig0&90.18&89.24&91.64&80.21&81.18&47.94\\
Sig1&91.80&91.02&93.14&84.03&77.96&58.35\\
Sig2&92.25&91.57&93.50&83.80&84.63&54.24\\
Sig3&92.35&91.70&93.57&84.37&82.88&59.81\\
\shline
Sig0  &92.59&91.86&93.91&83.22&86.16&78.61\\
Sig1&94.03&93.37&95.20&86.21&86.68&81.15\\
Sig2&94.37&93.82&95.44&86.62&90.15&81.28\\
Sig3&94.52&93.99&95.62&87.00&88.30&82.49\\
\whline
\end{tabular}
\begin{tablenotes}
    \footnotesize
    \item[1] The right columns list the CRs for different character types.
    \item[2] Upper: Dataset-ICDAR; Lower: Dataset-CASIA.
\end{tablenotes}
\end{threeparttable}
\end{table}

Table~\ref{Table:the effect of iterated integral of path signature} summarizes the recognition results of FCRN with path signature for different truncated versions (Sig0, Sig1, Sig2, and Sig3). Sig0 implies that only the $k=0$ iterated integral is considered in the experiments, Sig1 implies that the $k=0$ and $k=1$ iterated integrals are considered in the experiments, and so on for Sig2 and Sig3.
The experiments showed that the system performance improves monotonically from 90.18\% to 92.35\% on Dataset-ICDAR and from 92.59\% to 94.52\% on Dataset-CASIA as the path signature increases from Sig0 to Sig3. This proves the effectiveness of applying path signature to the OHCTR problem. 
Such results are obtained because the path signature captures more essential information from the pen-tip trajectories with higher iterated integrals.
We also observed that the performance improvement slows down as the iterated integrals increase, because the iterated integral of the path increases rapidly in dimension with severe computational burden while carrying very little information.
As Sig2 achieves a reasonable trade-off between efficiency and complexity, we selected it for feature extraction in the following experiments.

\begin{figure}[t]
\centering
\includegraphics[width=0.45\textwidth]{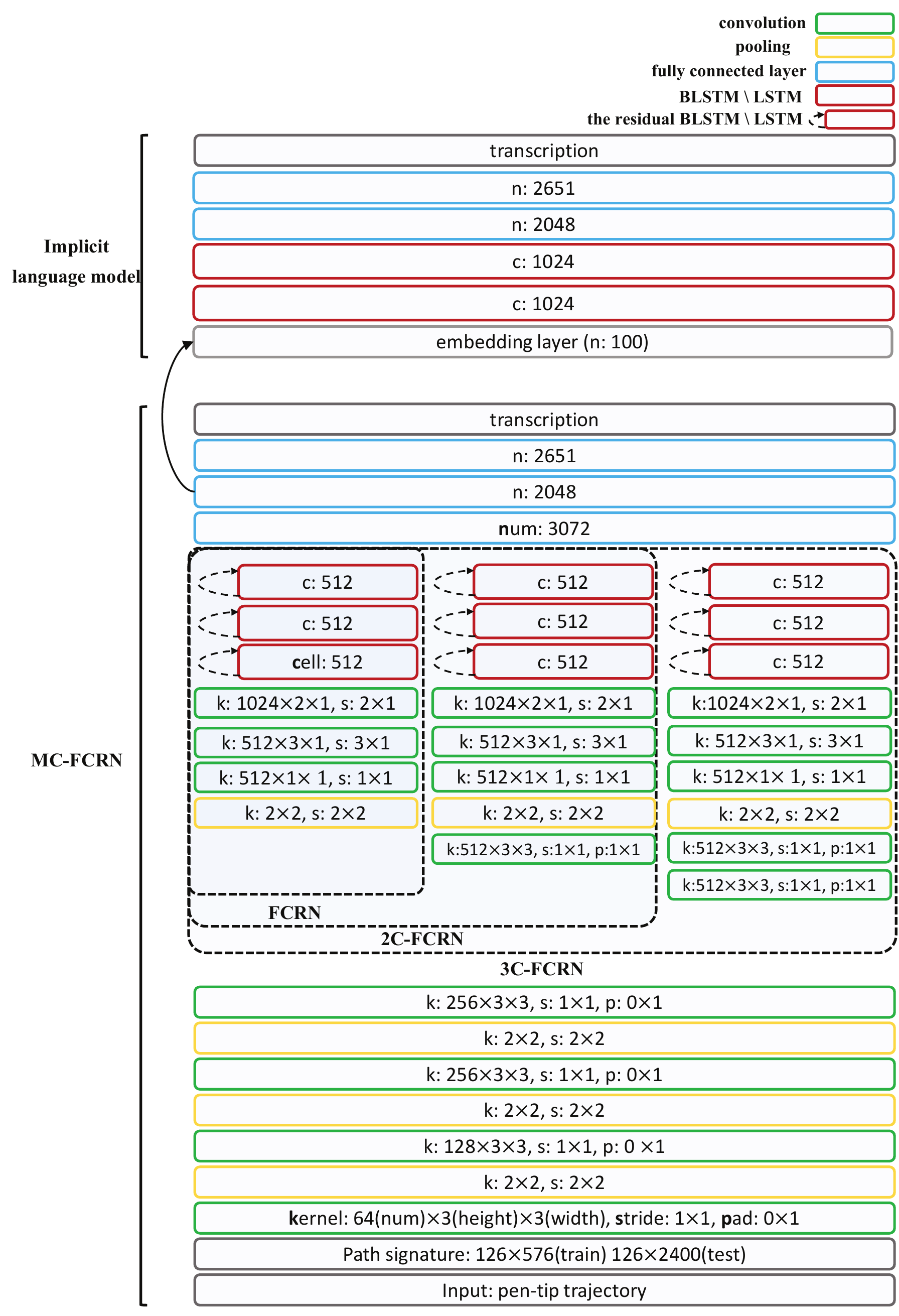}
\caption{Illustration of network architecture of MC-FCRN.}
\label{Fig:networkArchicture}
\end{figure}

\subsubsection{Effect of Spatial Context}
\label{sec:Effect of spatial context}
The effects of different spatial contexts are summarized in Table~\ref{Table:the effect of spatial context}, and the network architectures of FCRN, 2C-FCRN, and 3C-FCRN are shown in Fig.~\ref{Fig:networkArchicture}. From Fig.~\ref{Fig:networkArchicture}, we can see that FCRN, 2C-FCRN, and 3C-FCRN have one, two, and three receptive fields of different scales for each time step, respectively. The experiments showed that the system performance improved monotonically for both Dataset-ICDAR and Dataset-CASIA in the order of FCRN, 2C-FCRN, and 3C-FCRN, suggesting that we successfully leveraged the multiple spatial contexts by using multiple receptive fields and improved the system performance. Furthermore, we designed FCRN-2 and FCRN-3 such that their architectures and sizes were similar to those of 2C-FCRN and 3C-FCRN, except that their receptive fields for each time step were of the same scale. As shown in Table~\ref{Table:the effect of spatial context}, the recognition results of FCRN-2 and FCRN-3 were nearly the same as those of the original FCRN, which further verifies the improved performance of 2C-FCRN and 3C-FCRN from the additional spatial context.

\begin{table}[t]
\scriptsize
\caption{Effect of Spatial Context (Percent)}
\label{Table:the effect of spatial context}
\centering
\centering
\begin{threeparttable}
\begin{tabular}{cIccIcccc}
\whline
Network&CR&AR&\textsl{Chinese}&\textsl{Symbol}&\textsl{Digit}&\textsl{Letter}\\
\whline
FCRN&92.25&91.57&93.50&83.80&84.63&54.24\\
2C-FCRN&93.08&92.32&94.31&84.25&85.27&63.68\\
3C-FCRN&93.53&92.86&94.71&84.09&90.34&64.89\\
FCRN-2&92.40&91.35&93.47&84.16&88.81&65.38\\
FCRN-3&92.15&91.32&93.35&82.77&89.14&57.63\\
\shline
FCRN&94.37&93.82&95.44&86.62&90.15&81.28\\
2C-FCRN&94.94&94.34&96.07&86.82&90.09&83.16\\
3C-FCRN&95.16&94.58&96.22&86.78&93.18&87.30\\
FCRN-2&94.49&93.56&95.50&86.46&92.57&89.30\\
FCRN-3&94.17&93.49&95.29&85.65&90.62&84.76\\
\whline
\end{tabular}
\begin{tablenotes}
    \footnotesize
    \item[1] The right columns list the CRs for different character types.
    \item[2] Upper: Dataset-ICDAR; Lower: Dataset-CASIA.
\end{tablenotes}
\end{threeparttable}
\end{table}

\subsubsection{Effect of Semantic Context}
\label{sec:Effect of semantic context}

\newcolumntype{L}[1]{>{\raggedright\arraybackslash}p{#1}}  
\newcolumntype{C}[1]{>{\centering\arraybackslash}p{#1}}  
\newcolumntype{R}[1]{>{\raggedleft\arraybackslash}p{#1}}

\begin{table}[b]
\scriptsize
\caption{Effect of Semantic Context (Percent)}
\label{Table:the effect of language model}
\centering
\centering
\begin{threeparttable}
\begin{tabular}{cIC{0.45cm}C{0.45cm}IC{0.6cm}C{0.6cm}C{0.6cm}C{0.6cm}}
\whline
Network&CR&AR&\textsl{Chinese}&\textsl{Symbol}&\textsl{Digit}&\textsl{Letter}\\
\whline
3C-FCRN&93.53&92.86&94.71&84.09&90.34&64.89\\
3C-FCRN+S\_PFR&94.44&93.96&95.54&85.23&93.51&64.89\\
3C-FCRN+S\_PH &94.42&93.93&95.53&85.65&94.20&52.78\\
3C-FCRN+S\_SLD&94.59&94.15&95.60&85.41&94.52&78.69\\
3C-FCRN+B\_PFR&94.90&94.33&95.92&86.56&93.97&63.22\\
3C-FCRN+B\_PH&94.80&94.22&95.90&86.17&94.16&51.82\\
3C-FCRN+B\_SLD&95.19&94.66&96.18&86.32&94.25&79.90\\
3C-FCRN+B\_SLD+PFR&96.48&95.65&97.35&88.84&96.00&77.48\\
3C-FCRN+B\_SLD+PH&96.53&95.74&97.43&88.76&96.23&73.85\\
3C-FCRN+B\_SLD+SLD&97.15&96.50&97.92&90.10&96.78&86.68\\
\shline
3C-FCRN&95.50&94.73&96.46&88.55&91.62&85.43\\
3C-FCRN+PFR&96.63&95.98&97.46&89.85&96.74&82.75\\
3C-FCRN+PH&96.68&96.04&97.55&89.83&96.54&75.27\\
3C-FCRN+SLD&97.10&96.65&97.91&90.88&97.23&92.65\\
\whline
\end{tabular}
\begin{tablenotes}
    \footnotesize
    \item[1] The right columns list the CRs for different character types.
    \item[2] Upper: Dataset-ICDAR; Lower: Dataset-CASIA.
\end{tablenotes}
\end{threeparttable}
\end{table}

For implicit language model learning, we used the pre-trained 3C-FCRN as the recognizer. 
The networks denoted by 3C-FCRN+S\_PFR use single-directional LSTM for the implicit language model with synthesis training samples based on the PFR corpus, while the networks denoted by 3C-FCRN+B\_PFR use bidirectional LSTM (BLSTM) for the implicit language model.
As shown in Table~\ref{Table:the effect of language model}, the system performance was significantly improved with the implicit language model trained using the PFR, PH, and SLD corpora. Moreover, BLSTM-based networks performed much better than LSTM-based networks, suggesting that leveraging semantic context information from both forward and reverse directions of the text improves system performance.
In particular, the implicit language model that learns semantic context from SLD (3C-FCRN+B\_SLD) outperforms the other two models (3C-FCRN+B\_PFR and 3C-FCRN+B\_PH), especially on the \textsl{Letter} item. 
We also found that the system performance can be further improved by jointly applying the implicit language model and the statistical language model.
Actually, we applied the statistical language model to decode the result of  3C-FCRN+B\_SLD and the SLD corpus, i.e., 3C-FCRN+B\_SLD+SLD, again achieves superior performance with the best CRs on all character types. This advantage can be attributed to the size of its corpus (around 56 million characters), which is much greater than that of the PFR and PH corpora.

It is worth noting that it is unfair to apply the implicit language model to Dataset-CASIA because Dataset-CASIA has the same corpus as the training set of CASIA2.0-2.2. Thus, we directly applied the statistical language model to decode the prediction sequence of 3C-FCRN and observed a significant  improvement in performance on Dataset-CASIA.

\begin{table}[b]
\scriptsize
\newcommand{\tabincell}[2]{\begin{tabular}{@{}#1@{}}#2\end{tabular}}
\caption{Comparison with Previous Methods Based on Correct Rate and Accuracy Rate (Percent) for Dataset-ICDAR and Dataset-CASIA}
\label{TableFinalResult}
\centering
\begin{tabular}{cIcIcIcc}
\whline
Dataset&Methods&\tabincell{c}{language\\modeling}&CR&AR\\
\whline
\multirow{8}{*}{Dataset-ICDAR}
&CRNN \cite{shi2015end}&\XSolid&90.18&89.24\\
&3C-FCRN&\XSolid&93.53&92.86\\
&Zhou et al., 2013 \cite{zhou2013handwritten}&\Checkmark&94.62&94.06\\
&Zhou et al., 2014 \cite{zhou2014minimum}&\Checkmark&94.76&94.22\\
&VO-3 \cite{yin2013icdar}&\Checkmark&95.03&94.49 \\
&3C-FCRN+B\_SLD&\Checkmark&95.19&94.66\\
&3C-FCRN+B\_SLD+SLD&\Checkmark&\textbf{97.15}&\textbf{96.50}\\
\shline
\multirow{7}{*}{Dataset-CASIA}
&CRNN \cite{shi2015end}&\XSolid&92.59&91.86\\
&3C-FCRN&\XSolid&95.50&94.73\\
&Wang et al., 2012 \cite{wang2012approach}&\Checkmark&92.76&91.97 \\
&Zhou et al., 2013 \cite{zhou2013handwritten}&\Checkmark&94.34&93.75 \\
&Zhou et al., 2014 \cite{zhou2014minimum}&\Checkmark&95.32 & 94.69\\
&3C-FCRN+SLD&\Checkmark& \textbf{97.10}&\textbf{96.65} \\
&3C-FCRN+B\_CASIA&\Checkmark& \textbf{97.75}&\textbf{97.31} \\
\whline
\end{tabular}
\end{table}

\begin{figure*}%
\centering
\subfloat[]{
  \label{fig:ErrorAnalysis_a}
  \includegraphics[width=\textwidth]{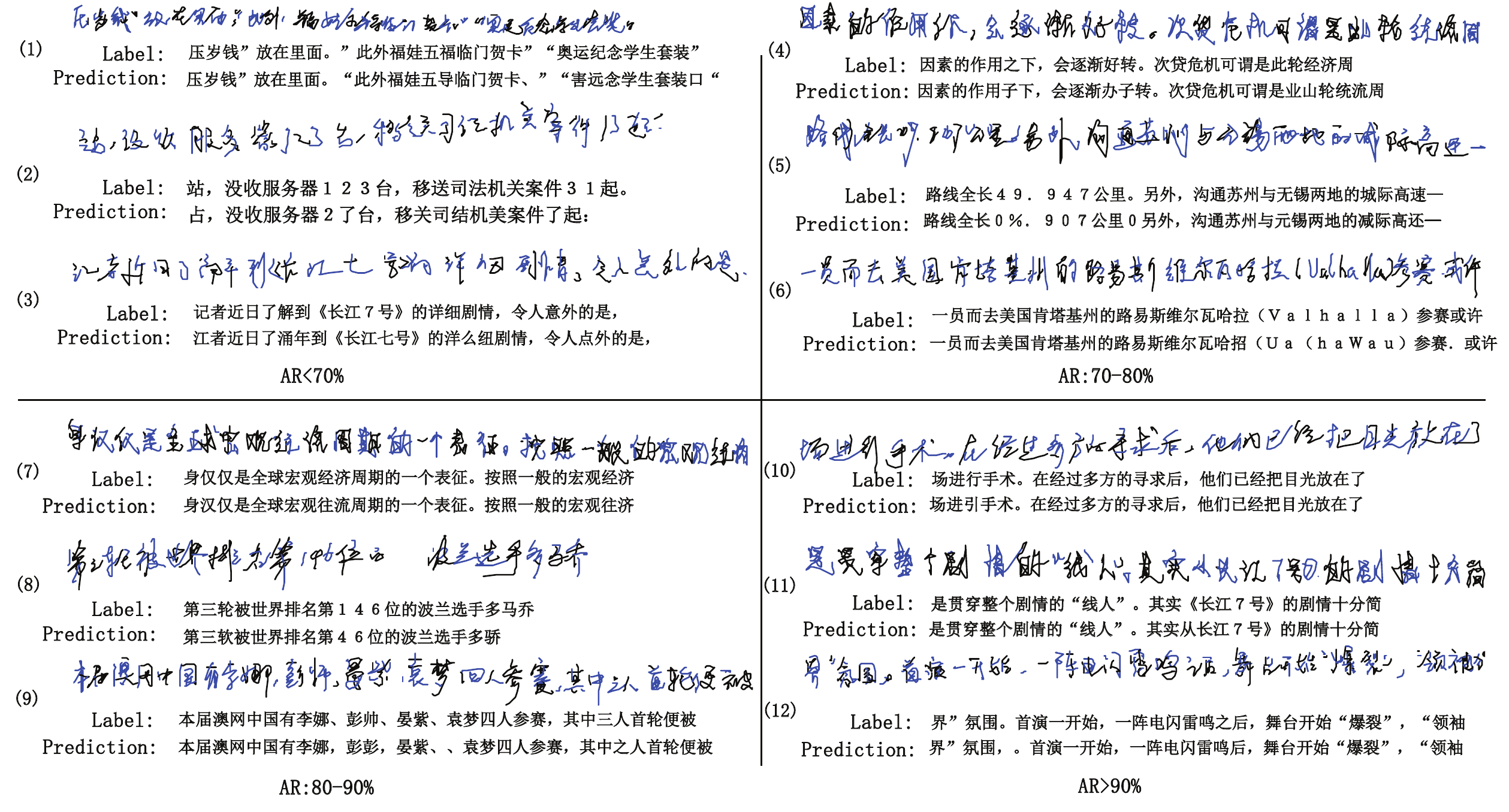}
  }\\
\subfloat[]{
  \label{fig:ErrorAnalysis_b}
  \includegraphics[width=0.18\textwidth]{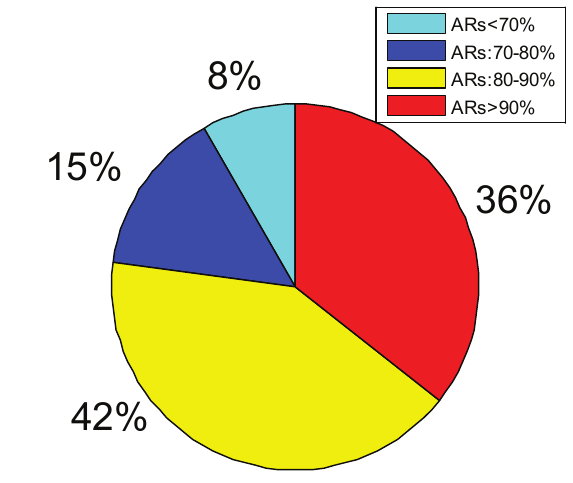}
  }\qquad
\subfloat[]{
  \label{fig:ErrorAnalysis_c}
  \includegraphics[width=0.75\textwidth]{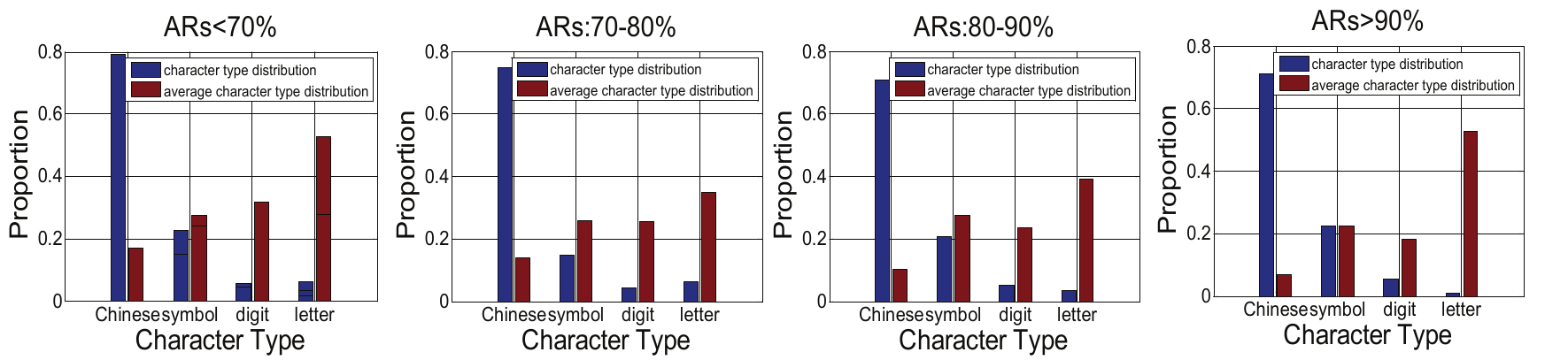}
  }
\caption{(a) Typical samples in different AR ranges. (b) Pie chart showing the proportion of different AR ranges. (c) Histogram showing character type distribution and average character type distribution for different AR ranges.}
\label{fig:ErrorAnalysis}
\end{figure*}
\subsubsection{Comparison with Previous Methods}
\label{sec:Comparison with previous methods}
The CRNN architecture proposed by Shi et al. \cite{shi2015end} is a special case of our MC-FCRN with only one receptive field in each time step and path signature truncated at zero (i.e., Sig0). As shown in Table~\ref{TableFinalResult}, we compared their network with our MC-FCRN without language modeling. Our MC-FCRN significantly outperformed CRNN on both Dataset-ICDAR and Dataset-CASIA, suggesting that MC-FCRN captures more essential spatial context information and online information from the pen-tip trajectories and is thus a better choice for the OHCTR problem.

The methods of Wang et al. \cite{wang2012approach}, Zhou et al. \cite{zhou2013handwritten}\cite{zhou2014minimum} and VO-3 \cite{yin2013icdar}, which are all based on the segmentation strategy, are completely different from our segmentation-free methods that incorporate the recently developed FCN, LSTM, and CTC.
On Dataset-ICDAR, our multi-spatial-context FCRN with the implicit language model (3C-FCRN+B\_SLD) outperformed all the previous methods. When further decoded with the trigram language model based on the SLD corpus, the results of our system (3C-FCRN+B\_SLD+SLD) were better than the best reported results from VO-3 \cite{yin2013icdar} with relative error reductions of 43\% and 36\% in CR and AR, respectively. Even when the outlier characters were directly considered to be completely wrong in our result, our system achieved outstanding performance with a CR of 95.13\% and an AR of 94.48\%, which is still the best result among all the reported results.
On Dataset-CASIA, 3C-FCRN+SLD outperformed all the other methods, with relative error reductions of 38\% and 37\% compared to the result of Zhou et al. \cite{zhou2014minimum} in terms of CR and AR, respectively.
The network denoted by 3C-FCRN+B\_CASIA use BLSTM for the implicit language model with the training set of CASIA2.0-2.2 (i.e., without synthesis samples based on corpora). 
As shown in Table~\ref{TableFinalResult}, 3C-FCRN+B\_CASIA outperform 3C-FCRN+SLD, which we claim to be an unfair comparison (see Section~\ref{sec:Effect of semantic context}) but somehow demonstrates the outstanding capability of our proposed implicit language model to incorporate semantic context in the recognition procedure.

\subsection{Error Analysis}
\label{sec:Error Analysis}
Fig.~\ref{fig:ErrorAnalysis} shows some quantitative analysis results for the erroneous recognized samples based on our 3C-FCRN. 
The proportion of samples in different AR ranges ($<70\%$, $70-80\%$, $80-90\%$, and $>90\%$) is displayed in the form of a pie chart (Fig.~\ref{fig:ErrorAnalysis_b}). Most of the erroneous recognized handwritten text lines fall in AR ranges of $80-90\%$ and $>90\%$ while only $8\%$ of the samples are recognized with AR lower than $70\%$. 
As shown in Fig.~\ref{fig:ErrorAnalysis_c}, for each AR range, we present the distribution of the erroneous recognized characters among different character types as well as the average distribution through division by the total number of character types.
The result is fairly consistent for different AR ranges, i.e., \textsl{Chinese} and \textsl{symbol} always account for the greater proportion in character type distribution, while the average character type distribution shows that it is quite difficult to recognize \textsl{symbol}, \textsl{digit}, and especially \textsl{letter} in OHCTR. 
We further investigated the erroneous recognized samples, some of which are shown in Fig.~\ref{fig:ErrorAnalysis_a}, to gain additional insights. The cursive nature of the unconstrained written samples is the main causes of the lower ARs, e.g., the small writing style and the character touching problem in sample 1, severe skew or slant text lines in sample 2, and ambiguous left-right structure in sample 8. 
Furthermore, it is difficult to recognize \textsl{digit} and \textsl{letter}, which lack cursive written samples for training (samples 2, 5, 6, and 8).
In addition, there are some similar characters (such as in samples 4, 7, 10, and 11) or even erroneous labeled characters (sample 3) in the test set that are very difficult to recognize.

\section{Conclusion}
\label{sec:conclusion}
In this paper, we addressed the challenging problem of unconstrained online handwritten Chinese text recognition by proposing a novel system that incorporates path signature, a multi-spatial-context fully convolutional recurrent network (MC-FCRN), and an implicit language model.
We exploited the spatial structure and online information of online pen-tip trajectories with a powerful path signature by using a sliding window-based method. Experiments showed that the path signature truncated at level two achieves a reasonable trade-off between efficiency and complexity for OHCTR.
For spatial context learning, we demonstrated that our MC-FCRN successfully exploits multiple spatial contexts from receptive fields with multiple scales to robustly recognize the input signature feature maps.
For semantic context learning, an implicit language model was developed to learn to make predictions conditioned on the entire predicting feature sequence, significantly improving the system performance, especially when combined with the statistical language model. 
In the experiments, our best result significantly outperformed all other existing methods, with relative error reductions of 43\% and 38\% in terms of the correct rate on two standard benchmarks, Dataset-ICDAR and Dataset-CASIA, respectively.

In the future, we plan to incorporate the over-segmentation strategy as prior knowledge in our network enhance the recognition performance. Furthermore, in this paper, our character set was limited to only 2650 classes with the training set of CASIA2.0-2.2. How to deal with the out-of-vocabulary issue is still a challenging problem to be investigated in the future.

\section*{Acknowledgment}
TL and HN are supported by the Alan Turing Institute under the EPSRC grant EP/N510129/1. TL is also supported by ERC advanced grant ESig (agreement no. 291244).

\bibliographystyle{IEEEtran}
\bibliography{refs}

\end{document}